\documentclass[journal]{IEEEtran}
\pdfoutput=1

\usepackage{graphicx,times,multirow,cite} 
\usepackage{algorithm}
\usepackage{clrscode}
\usepackage{algpseudocode}
\usepackage{enumerate}
\usepackage{caption}
\usepackage{longtable}
\usepackage{array}
\usepackage{amsmath,amssymb}
\usepackage{pst-blur}
\usepackage{bm}
\usepackage{pstricks-add}
\usepackage{changepage}
\usepackage{subfigure}
\usepackage{epstopdf}
\usepackage{fancyhdr} 
\definecolor{Blue}{rgb}{1.0,0.75,0.8}
\newcommand{\qed}{\nobreak \ifvmode \relax \else
      \ifdim\lastskip<1.5em \hskip-\lastskip
      \hskip1.5em plus0em minus0.5em \fi \nobreak
      \vrule height0.75em width0.5em depth0.25em\fi}

\DeclareGraphicsExtensions{.pdf,.png,.jpg,.eps,.fig}
\graphicspath{{images/}}

\ifCLASSINFOpdf
\else
\fi
\begin{document}
\title{Negatively Correlated Search}
\author{Ke Tang,\textit{ Senior Member, IEEE}, Peng~Yang,\textit{ Student Member, IEEE} and Xin~Yao,\textit{ Fellow, IEEE} 
\thanks{This work was supported in part by the National Natural Science Foundation of China (Grant Nos. 61329302 and 61175065), the Program for New Century Excellent Talents in University (Grant No. NCET-12-0512), EPSRC (Grant No. EP/J017515/1) and a Royal Society Newton Advanced Fellowship (Ref. No. NA150123). Xin Yao was also supported by a Royal Society Wolfson Research Merit Award.

Ke Tang and Peng Yang are with the USTC-Birmingham Joint Research Institute in Intelligent Computation and Its Applications, School of Computer Science and Technology, University of Science and Technology of China (USTC), Hefei, Anhui 230027, China (e-mail: ketang@ustc.edu.cn; trevor@mail.ustc.edu.cn). Corresponding Author: Ke Tang.

Xin Yao is with the Centre of Excellence for Research in Computational
Intelligence and Applications (CERCIA), School of Computer Science, The
University of Birmingham, Edgbaston, Birmingham B15 2TT, UK (e-mail:
x.yao@cs.bham.ac.uk)}
}
\markboth{}%
{Shell \MakeLowercase{\textit{et al.}}: Bare Demo of IEEEtran.cls for Journals}
\maketitle
\begin{abstract}
Evolutionary Algorithms (EAs) have been shown to be powerful tools for complex optimization problems, which are ubiquitous in both communication and big data analytics. This paper presents a new EA, namely Negatively Correlated Search (NCS), which maintains multiple individual search processes in parallel and models the search behaviors of individual search processes as probability distributions. NCS explicitly promotes negatively correlated search behaviors by encouraging differences among the probability distributions (search behaviors). By this means, individual search processes share information and cooperate with each other to search diverse regions of a search space, which makes NCS a promising method for non-convex optimization. The cooperation scheme of NCS could also be regarded as a novel diversity preservation scheme that, different from other existing schemes, directly promotes diversity at the level of search behaviors rather than merely trying to maintain diversity among candidate solutions. Empirical studies showed that NCS is competitive to well-established search methods in the sense that NCS achieved the best overall performance on 20 multimodal (non-convex) continuous optimization problems. The advantages of NCS over state-of-the-art approaches are also demonstrated with a case study on the synthesis of unequally spaced linear antenna arrays.
\end{abstract}
\begin{IEEEkeywords}
 Evolutionary Algorithms, Negative Correlation, Diversity Maintenance, Optimization in Communication Systems
\end{IEEEkeywords}
\IEEEpeerreviewmaketitle

\section{Introduction}
\IEEEPARstart{T}{he} design or application of a communication system often requires solving a challenging optimization problem. For example, minimizing the Symbol-Error-Rate (SER) for Amplify-and-Forward Relaying Systems is non-trivial since the SER surface is non-convex and has multiple minima \cite{Ahmed2014}. When tuning the protocol of sensornets \cite{Tate2012}, as mathematical modeling of sensornets involves numerous inherent difficulties, one might has to deal with the optimization problem without an explicit objective function and the quality of candidate protocol configurations could only be obtained from a simulation model. In the era of Big Data, as data analytics (e.g., machine learning) fast grows into a ubiquitous technology in many areas, including communications, the challenges brought by complex optimization problems become even more important than ever. First, one of the major roles of big data analytics is to acquire knowledge from data in order to facilitate decision-making, e.g., managing network resources based on the analysis of user profiles to achieve higher end-users satisfaction\cite{mushtaq2014qoe}. Thus, the value of big data usually needs to be created through tackling an optimization problem (i.e., to seek the optimal decision), which is formulated based on the knowledge obtained from big data analytics. Such optimization problems may not only be non-convex, but also be noisy due to the noise contained in the original data. Furthermore, the data analytics process may also involve complex optimization problems, such as training deep neural networks \cite{david2014genetic} or tuning the hyper-parameter of Support Vector Machines \cite{koch2012tuning}. To cope with these complex optimization problems, Evolutionary Algorithms (EAs) have been widely adopted and been shown to be a family of powerful tools \cite{david2014genetic,wang2015convex,qian2015pareto}.
In short, to create business values from big data analytics, optimization tools are indispensable. 

EAs search in the solution space of a problem by iteratively generating a population of new candidate solutions until a predefined halting condition is met. The use of population (rather than generating a single solution at each iteration) has been proved to be critical to the success of EAs both theoretically \cite{he2002individual} and empirically \cite{mallipeddi2008empirical}. In particular, it is widely believed that information should be shared among individual solutions in the population, so that they can cooperatively generate new, hopefully more promising, candidate solutions and the algorithm could eventually search more effectively. The key research issues here are what information to share and how. In the literature, such a design philosophy is mainly implemented either by directly combining two or more candidate solutions to generate new ones using the so-called reproduction operators \cite{back1996evolutionary}, or by employing the population to build probabilistic models for generating new solutions \cite{larranaga2002estimation}. 

In this paper, we propose a new population-based search method, namely Negatively Correlated Search (NCS). The core idea of NCS is a new model for implementing the cooperation between individuals in a population, which was inspired by an interpretation of cooperation in human behaviors. That is, when a team of people is tackling a complex task, members of the team tend to work cooperatively by handling different parts of the task and communicate to avoid multiple members working on the same part. Analogously, NCS comprises multiple search processes. The search processes are run in parallel and strive to find better candidate solutions, while information is shared to explicitly encourage each search process to emphasize the regions that are not covered by others. 

In Section \uppercase\expandafter{\romannumeral2}, we will describe the general idea of NCS. Since the core idea of NCS is closely related to another important research issue, namely diversity maintenance, of EAs, Section \uppercase\expandafter{\romannumeral3} will further discuss the novelty of NCS in this aspect by comparing its core idea with existing diversification schemes. After that, a concrete implementation of NCS for continuous optimization problems, namely NCS-C, will be detailed in Section \uppercase\expandafter{\romannumeral4}. In Section \uppercase\expandafter{\romannumeral5}, the advantages of NCS-C will be demonstrated through comparisons with other stochastic search methods on challenging benchmark problems. Section \uppercase\expandafter{\romannumeral6} further demonstrates the potential of NCS on a real-world problem, namely the Synthesis of Unequally Spaced Antenna Arrays (SUSAA). Finally, conclusions will be drawn in Section \uppercase\expandafter{\romannumeral7}.

\section{General Framework of Negatively Correlated Search}
We consider a population-based search process as multiple iterative search processes that are run in parallel. Each iterative search process basically consists of 3 procedures. First, one or more initial solutions are created randomly. Then, a randomized search operator, e.g., the crossover or mutation operator of a Genetic Algorithm (GA), is applied to the current solutions to iteratively generate new candidate solutions. Finally, the search terminates when a predefined criterion is met. Many existing population-based search method can be formulated in this way as long as there exists some kind of relationship between solutions generated in different iterations.

Intuitively, any population-based search method captured by the above procedures could be viewed as a multi-agent system \cite{ferber1999multi}, in which every agent (i.e., a search process) strives to find a solution with better quality in terms of the value of the objective function of an optimization problem. Cooperation between the agents is expected to facilitate the search. Various strategies could contribute to designing a cooperation scheme for this purpose. The one adopted in NCS is that different agents should cover different regions of the search space and have different search behaviours. Specifically, NCS expects each search process to move towards a region that is both promising and is unlikely to be searched by other search processes, as elaborated below.

For the sake of simplicity, we consider a special case that each search process is a Randomized Local Search (RLS), which produces one new candidate solution in each iteration. In other words, we consider a population-based search method with population size of $N (N>1)$, which runs $N$ RLS procedures in parallel and iteratively updates each individual solution in the population with a randomized search operator. Suppose at the $t$th iteration of NCS, $N$ new solutions have been generated by applying some randomized search operators to $N$ current solutions, respectively. Since randomized search operators are employed to generate new solutions, the search bias of a RLS in the $t+1$th iteration is essentially a probability distribution from which a new solution will be sampled. Hence, if the probability distribution corresponding to a RLS differs from those corresponding to the other RLSs, it is likely to generate solutions in a region that is less likely to be covered by other RLSs. This requires measuring the difference or correlation between two probability distributions, for which the Bhattacharyya Distance \cite{Kailath1089532} can be employed. Eqs. (1) and (2) gives the Bhattacharyya distance for two continuous and discrete probability distributions, respectively:

\begin{eqnarray} 
   D_B(p_i,p_j)=-\ln\bigg(\int{\sqrt{p_i(\textbf{x})p_j(\textbf{x})}}\text{d} \textbf{x}\bigg)   \\
   D_B(p_i,p_j)=-\ln\bigg(\sum_{\textbf{x}\in\textbf{X}}{\sqrt{p_i(\textbf{x})p_j(\textbf{x})}}\bigg) 
\end{eqnarray}

\noindent where $p_i$ and $p_j$ denote the probability density functions of two distributions. In case that the probability density function is not explicitly known, one may randomly generate a set of candidate solutions for both distributions and estimate the Bhattacharyya distance using the Bhattacharyya coefficient.

Let $\textbf{x}_i$ denote the current solution obtained by the $i$th RLS and $\textbf{x}_i'$ denote the new solution generated based on $\textbf{x}_i$. At each iteration, one of the two solutions will be selected for generating another new solution in the next iteration and the other will be discarded. The probability distribution associated with the ith RLS in the next iteration depends on the search operator and the solution kept, i.e., either $\textbf{x}_i$ or $\textbf{x}_i'$. Hence, by choosing a solution to keep, the ith RLS actually chooses one of the corresponding distributions, denoted as $p_i$ and $p_i'$. Ideally, the selected solution should be with high quality and should lead to a distribution that is distant from those corresponding to the other RLSs. In NCS, the former can be measured with the value of the objective function to be minimized, denoted as $f(\textbf{x})$. The latter is defined as:

\begin{eqnarray} 
Corr(p_i) =  \underset{j}{\min}\{D_B(p_i,p_j)\vert{j\neq i}\}
\end{eqnarray}

\noindent where $p_j$ represents the distributions corresponding to other RLSs and a larger $Corr(p_i)$ is preferred. 
 
Determining an appropriate trade-off for the above cases can be difficult because $f(\textbf{x})$ and $Corr(p)$ may be of different scales and $f(\textbf{x})$ may take negative values while $Corr(p)$ is always non-negative. Hence, non-negative objective function values need to be first guaranteed by subtracting the minimum objective function value achieved so far from $f(\textbf{x}_i)$ and $f(\textbf{x}_i')$. Then, normalization is conducted by requiring $f(\textbf{x}_i )+f(\textbf{x}_i')$ and $Corr(p_i)+Corr(p_i')$ equal to 1. With the normalization step, it is no longer necessary to involve $f(\textbf{x}_i)$ and $Corr(p_i)$ for deciding which solution to discard because they now equal to $1-f(\textbf{x}_i')$ and $1-Corr(p_i')$, respectively. The smaller the $f(\textbf{x}_i')$, the better $\textbf{x}_i'$ is in the sense of solution quality (assuming we are solving minimization problems). The larger the $Corr(p_i')$, the less likely that $\textbf{x}_i'$ will result in new solutions that are close to the solutions generated by other RLSs. Thus, NCS prefers the solution $\textbf{x}_i'$ associated with small $f(\textbf{x}_i')$ and large $Corr(p_i')$, and the following heuristic rule is adopted in NCS to integrate the solution quality and correlation between RLSs into one selection criterion:

\begin{eqnarray} 
 \left\{ 
\begin{array}{ll}  
\text{discard} \,\, \textbf{x}_i,& {\textbf{if} \quad \frac{f(\textbf{x}_i')}{Corr(p_i')}<\lambda}\\
\text{discard} \,\, \textbf{x}_i',& {\textbf{otherwise}} 
\end{array}
\right.
\end{eqnarray}

 \noindent where $\lambda \in (0, +\infty)$ is a parameter. 

To summarize, NCS maintains a population of RLSs and iteratively generates new solutions. At each iteration, each RLS is employed to generate one new solution. The new solution is then compared with the solution generated by the same RLS in the previous iteration. One of the two solutions will be discarded according to Eq. (4). The search terminates when a predefined halting condition is satisfied.

\section{NCS as a Diversity Maintenance Approach}

The search mechanism of NCS can be made clearer by putting it into the general background of EAs, where the trade-off between exploration and exploitation is probably the most widely discussed issue \cite{vcrepinvsek2013exploration}. Specifically, most EAs maintain the diversity of a population to control the exploration power, so as to promote search in new promising regions and avoid premature convergence \cite{vcrepinvsek2013exploration}. Existing strategies for diversity maintenance can be categorized into two types: (1) The niching techniques \cite{das2011real}, such as fitness sharing and crowding, aim to select a set of solutions that are distant from one another in the solution space. These solutions will then be utilized to generate new solutions. This type of methods neglects the fact that a set of diverse (distant) solutions does not necessarily produce a diverse set of new solutions, since the latter also depends on the search operators used. (2) Alternatively, search operators with a large search step size could be employed periodically to generate new solutions that are likely to be more distant to the current solutions \cite{yao1999evolutionary}. In the extreme case, the whole population could be randomly initialized \cite{jansen2002analysis}. However, this type of methods will introduce another algorithm design problem that is difficult to address in practice. That is, which (large step-size) search operators should be used and when to switch between operators with different search step-sizes during the search process to achieve a good trade-off between exploration and exploitation.

The core idea of NCS could be viewed as a novel diversity maintenance scheme. However, the diversity in the NCS framework is very different from the diversity in other cases, e.g., niching. Almost all other work defines diversity as diversity among individual solutions, e.g., genotypic diversity, while NCS defines diversity in terms of search behaviors, i.e., the probability distributions associated with RLSs. It is the search behaviour, which captures the on-going interaction between search operators and solutions, that really matters, not how a population of solutions "looks" (i.e., genotypic diversity). In certain  sense, the concept of diversity in NCS bears some similarity to that in negative correlation learning \cite{liu1999ensemble}. In this way, NCS explicitly prefers solutions that are likely to produce diverse good solutions, although the selected solutions are not necessarily distant from one another in the solution space. Since the ultimate goal of diversity maintenance is to promote search in new promising regions, NCS is more directly related to this goal and thus is expected to facilitate the search better. Moreover, even if a fixed search operator is used throughout the search process, NCS is capable of “pushing” a RLS to visit regions that are seldom covered by other RLSs. Hence, NCS does not rely on fine-tuned coordination of different search operators to maintain diversity.

\section{An instantiation of NCS for Continuous Optimization}	
	
In this section, an instantiation of NCS for continuous optimization is presented to illustrate the detailed steps of a NCS algorithm.

\subsection{Solution Representation and Search Operators}

Suppose a $D$-dimensional continuous minimization problem with objective function $f(\textbf{x})$ is to be tackled. In NSC-C, a solution is represented as a $D$-dimensional real-valued vector. The Gaussian mutation operator \cite{beyer2002evolution} is employed as the search operator for all RLSs in NCS-C. Given an existing solution $\textbf{x}_i$, the Gaussian mutation operator generates a new solution $\textbf{x}_i'$ using Eq. (5):

\begin{eqnarray} 
x_{id}'=x_{id}+\mathcal{N}(0,\sigma_i)
\end{eqnarray}

\noindent where $x_{id}$ denotes the $d$th element of $\textbf{x}_i$ and $\mathcal{N}(0,\sigma_i)$ denotes a Gaussian random variable with zero mean and standard deviation $\sigma_i$. In general, the value of $\sigma_i$ can be adapted during the search and may also vary over RLSs or even dimensions. To keep the algorithm simple, all RLSs in NCS-C are initialized with the same value of $\sigma_i$. Then, each $\sigma_i$ is adapted for every $epoch$ iterations according to the 1/5 successful rule suggested in \cite{beyer2002evolution}, as given in Eq. (6):

\begin{eqnarray} 
\sigma_i  = \left\{ 
\begin{array}{ll}  
 \frac{\sigma_i}{r} & \textbf{if} \quad \frac{c}{epoch} > 0.2\\ 
 \sigma_i*r &  \textbf{if} \quad \frac{c}{epoch} < 0.2\\ 
 \sigma_i  &  \textbf{if} \quad \frac{c}{epoch} = 0.2
\end{array} 
\right. 
\end{eqnarray}

\noindent where $r$ is a parameter that is suggested to be set beneath 1, and $c$ is the times that a replacement happens (i.e., $\textbf{x}_i'$ is preserved) during the past $epoch$ iterations. Eq. (6) is designed based on the following intuition. A large $c$ implies that the RLS frequently found better solutions in the past iterations, and the current best solution might be close to the global optimum. Thus, the search step-size should be reduced (by $r$ times). On the other hand, if a RLS frequently failed to achieve a better solution in the past iterations, it might have been stuck in a local optimum. In this case, the search step-size will be increased (by $r$ times) to help the RLS explore other promising regions in the search space.

According to Eq. (5), the Gaussian mutation operator generates a new solution based on each $\textbf{x}_i$ following a multivariate normal distribution. The expectation of the distribution is $\textbf{x}_i$ and the covariance matrix \boldsymbol{$\Sigma_i$} is $\sigma_i^2 \textbf{I}$, where $\textbf{I}$ is the identity matrix of size $D$. Hence, given two solutions $\textbf{x}_i$ and $\textbf{x}_j$, the Bhattacharyya distance given by Eq. (1) can be written as Eq. (7) \cite{fukunaga2013introduction} (page 99, Eq. 3.152):

\begin{eqnarray} 
\begin{array}{ll} 
D_B(p_i,p_j) =&\frac{1}{8}(\textbf{x}_i-\textbf{x}_j)^T \boldsymbol{\Sigma}^{-1}(\textbf{x}_i-\textbf{x}_j)\\
&+\frac{1}{2}\ln\bigg(\frac{\det\boldsymbol{\Sigma}}{\sqrt{\det\boldsymbol{\Sigma_i}\det\boldsymbol{\Sigma_j}}}\bigg) 
\end{array} 
\end{eqnarray}

where \boldsymbol{$\Sigma$} = $\frac{\boldsymbol{\Sigma_i}+\boldsymbol{\Sigma_j}}{2}$.

Subsequently, the criterion given in Eq. (4) can be computed by considering the Bhattacharyya distance between $\textbf{x}'_i$ and all the $\textbf{x}_j$ generated by the other RLSs.

\subsection{Adaptation of Parameter $\lambda$}

Given $\textbf{x}_i$ and $\textbf{x}_i'$, Eq. (4) shows that different values of $\lambda$ may lead to different decisions on which solution will be kept/discarded. Thus the value of $\lambda$ can affect the search process and consequently the performance of NCS. Setting $\lambda$ to 1 indicates that the solution quality and correlation between RLSs are equally emphasized and hence can be used as the default setting. However, different values of $\lambda$ might be suitable at different stages of the search. Hence, a time-variant $\lambda$ is employed in NCS-C. Specifically, at the late stage of the search, each RLS tends to shrink its search step size, and thus $\textbf{x}_i$ and $\textbf{x}_i'$ may take similar values of $f(\textbf{x})$ and $Corr(p)$. In this case, Eq. (4) is highly sensitive to $\lambda$ since $\frac{f(\textbf{x}_i')}{Corr(p_i')}$ is converging towards 1. In contrast, the difference between $\textbf{x}_i$ and $\textbf{x}_i'$ in terms of $\frac{f(\textbf{x}_i')}{Corr(p_i')}$ is relatively larger in the early stage of the search. Accordingly, a $\lambda$ away from the default setting might be beneficial. For these considerations, NCS-C randomly samples a Gaussian distribution with expectation 1 to generate values of $\lambda$ at each iteration. The standard deviation of the Gaussian distribution is first initialized to 0.1, and then shrink towards 0 according to Eq. (8) 
\begin{eqnarray}
\lambda_t=\mathcal{N}(1, 0.1-0.1*\frac{t}{T_{max}}) 
\end{eqnarray}

\noindent where $T_{max}$ is the user-defined total number of iterations for an execution of NCS-C.

Algorithm 1\footnote{The Matlab code of NCS-C can be downloaded at: http://staff.ustc.edu.cn/$\sim$ketang/codes/NCS.html} outlines the pseudo-code of NCS-C. A population of solutions will be randomly initialized and evaluated at lines 1-2. Then, the algorithm will iteratively repeat lines 5-25 until $T_{max}$ iterations have been conducted. At each iteration, the control parameter $\lambda$ in Eq. (4) is first set according to Eq. (8) at line 6. Then, new solutions $\textbf{x}'$s are generated by applying Eq. (5) to $\textbf{x}$ and evaluated with respect to both objective function $f$ and correlation $Corr$ according to Eqs. (3) and (7), as described at lines 7-10. The best solution found will be updated at lines 12-14. Since $f$ and $Corr$ may be of different scales, the two terms are normalized by requiring $f(\textbf{x}_i)+f(\textbf{x}_i')$ and $Corr(p_i)+Corr(p_i')$ equal to 1 before line 15. Lines 15-17 decide whether to replace $\textbf{x}$ with $\textbf{x}'$ based on Eq. (4). For every $epoch$ iteration, lines 20-24 update the search step-size for each RLS based on Eq. (6). Finally, the best solution found during the entire search process will be output when the search terminates at line 26.

      \begin{algorithm}[h] 
        \caption{NCS-C( $T_{max}$, \boldsymbol{$\sigma$}, $r$, $epoch$, $N$)} 
        \label{NCS-C} 
        \begin{algorithmic}[1] 
        \State Randomly generate an initial population of $N$ solutions.  
        \State Evaluate the $N$ solutions with respect to the objective function $f$.  
        \State Identify the best solution \textbf{x*} in the initial population and store it in \textit{BestFound}. 
	   \State Set $t \leftarrow 0$
        \State \textbf{While} ($t<T_{max}$) \textbf{do} 
            \State \quad   Set $\lambda_t \leftarrow \mathcal{N}(1, 0.1-0.1*\frac{t}{T_{max}})$.   
            \State \quad  \textbf{For} $i = 1$ \textbf{to} $N$   
            \State  \quad\quad  Generate a new solution $\textbf{x}_i'$ by applying Gaussian mutation operator with $\sigma_i$ to $\textbf{x}_i$.  
		 \State  \quad\quad  Compute $f(\textbf{x}_i'), Corr(p_i)$ and $Corr(p_i')$.
		 \State  \quad  \textbf{EndFor}
		 \State  \quad  \textbf{For} $i = 1$ \textbf{to} $N$ 
            \State  \quad\quad  \textbf{If}  $f(\textbf{x}_i')<f(\textbf{x*})$
            \State  \quad\quad\quad  Update \textit{BestFound} with $\textbf{x}_i'$.
            \State  \quad\quad  \textbf{EndIf}
		\State   \quad\quad \textbf{If}  $\frac{f(\textbf{x}_i')}{Corr(p_i')}<\lambda_t$
            \State  \quad\quad\quad  Update $\textbf{x}_i$ with $\textbf{x}_i'$.
            \State   \quad\quad \textbf{EndIf}
            \State   \quad \textbf{EndFor} 
            \State \quad   $t \leftarrow t + 1$ 
            \State  \quad  \textbf{If} mod($t, epoch$) = 0
            \State  \quad\quad  \textbf{For} $i = 1$ \textbf{to} $N$ 
            \State   \quad\quad\quad Update $\sigma_i$ for each RLS according to the 1/5 successful rule.
            \State  \quad\quad  \textbf{EndFor} 
            \State \quad   \textbf{EndIf}
        \State \textbf{EndWhile}  
	   \State  \textbf{Output} \textit{BestFound}
        \end{algorithmic} 
      \end{algorithm}

\begin{figure}\renewcommand{\captionfont}{\footnotesize}
			 \centering
			\begin{minipage}[htb]{1\linewidth}
				\centering
				\includegraphics[width=1\textwidth,height=0.75\textwidth]{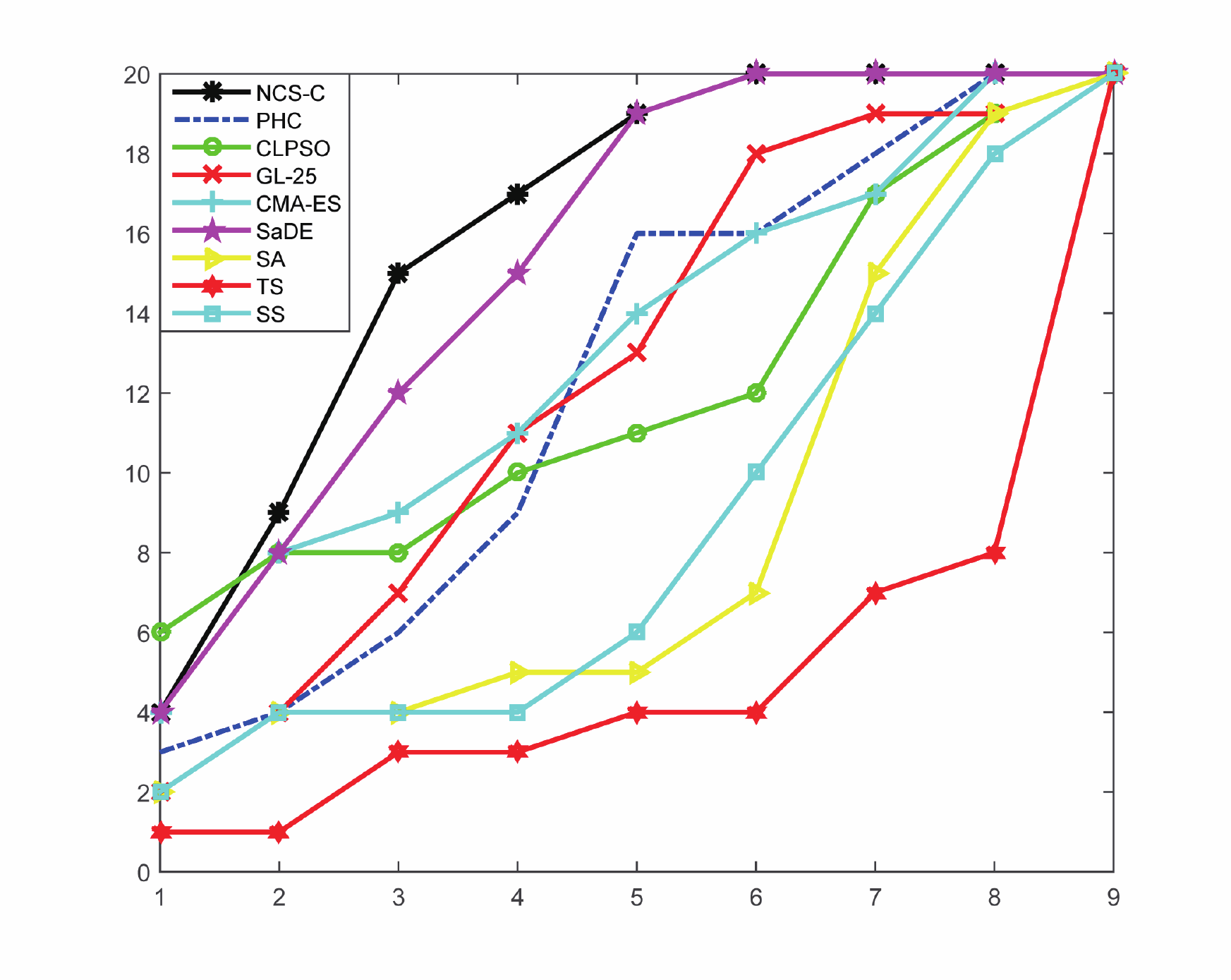}
				\caption{The Top-K, K =1,2,…,9, curves of the algorithms.}
			\end{minipage}%
\end{figure}

\section{Computational Studies}
To assess the potential of NCS, computational studies have been carried out to compare the NCS-C against a number of well-established population-based search methods, including GA, Particle Swarm Optimizer (PSO) \cite{kennedy2010particle}, Evolution Strategies (ES) \cite{beyer2002evolution}, Differential Evolution (DE)\cite{storn1997differential} and Scatter Search (SS) \cite{marti2006principles}. For each of these algorithms, a corresponding state-of-the-art version is chosen for comparison, i.e., GL-25 \cite{garcia2008global}, CLPSO \cite{liang2006comprehensive}, CMA-ES \cite{hansen2001completely}, SaDE \cite{qin2009differential} and the SS \cite{marti2006principles}, respectively. Furthermore, two well-known individual-based search methods, Simulated Annealing (SA) \cite{corana1987minimizing} and Tabu Search (TS) \cite{chelouah2000tabu}, are also included because of their appealing performance in practice. Besides, a variant of NCS-C that omits the negative correlation between RLSs, denoted as Parallel Hill-Climbing (PHC), is created in order to verify the impact of the negative correlations between RLSs on the NCS-C. The parameters of the above-mentioned compared algorithms were set following suggestions in the literature, as detailed below.

{\renewcommand\baselinestretch{0.8}\selectfont
\begin{table*}[!tbp]\scriptsize
\centering  
\caption{The averaged results of 9 algorithms on 20 benchmark multi-modal problems are listed in the form of 'mean $\pm$ standard deviation'. All the results are presented in terms of function errors, i.e., the difference between the objective function value of the obtained solution and that of the optimal solution to the problem. The last two rows provide the results of the Friedman test and Wilcoxon test, where 'w-d-l' indicates NCS-C is superior to, not significantly different from or inferior to the corresponding compared algorithms.}
\begin{tabular}{c||r|r|r|r|r|r|r|r|r}
\hline
\textbf{Algo.} &\textbf{PHC} &\textbf{SA} &\textbf{TS} &\textbf{SS} &\textbf{GL-25} &\textbf{SaDE} &\textbf{CMA-ES} &\textbf{CLPSO} &\textbf{NCS-C}\\ \hline

\multirow{2}*{\boldsymbol{$F_6$}} &2.61E+01 &3.90E+02 &7.00E+03 &2.17E+05 &2.13E+01 &4.76E+01 &\textbf{0.00E+00} &4.80E+00 &2.08E+01  \\ 
&$\pm$2.35E+01 &$\pm$4.09E+01 &$\pm$1.01E+04 &$\pm$6.41E+04 &$\pm$1.02E+01 &$\pm$3.35E+01 &$\pm$\textbf{0.00E+00} &$\pm$3.55E+00 &$\pm$3.61E+00 \\ \hline

\multirow{2}*{\boldsymbol{$F_7$}} &\textbf{9.86E-04} &2.21E+00 &1.64E-02 &1.40E+00 &2.78E-02 &1.95E-02 &1.84E-03 &4.63E-01 &1.69E-02  \\ 
&$\pm$\textbf{2.76E-03} &$\pm$1.84E+00 &$\pm$1.59E-02 &$\pm$7.31E-02 &$\pm$3.76E-02 &$\pm$1.37E-02 &$\pm$4.59E-03 &$\pm$7.31E-02 &$\pm$1.38E-02 \\ \hline

\multirow{2}*{\boldsymbol{$F_8$}} &\textbf{2.00E+01} &2.10E+01 &2.01E+01 &2.09E+01 &2.10E+01 &2.09E+01 &2.03E+01 &2.10E+01 &\textbf{2.00E+01}  \\ 
&$\pm$\textbf{1.29E-02} &$\pm$7.13E-02 &$\pm$3.60E-02 &$\pm$3.60E-02 &$\pm$5.12E-02 &$\pm$4.76E-02 &$\pm$5.62E-02 &$\pm$5.60E-02 &$\pm$\textbf{1.22E-02} \\ \hline

\multirow{2}*{\boldsymbol{$F_9$}} &1.07E+02 &2.41E+02 &4.83E+02 &2.57E+02 &2.63E+01 &1.99E-01 &4.12E+02 &\textbf{0.00E+00} &9.36E+01  \\ 
&$\pm$2.13E+01 &$\pm$8.62E+01 &$\pm$9.60E+01 &$\pm$3.85E+01 &$\pm$5.64E+00 &$\pm$4.06E-01 &$\pm$1.38E+02 &$\pm$\textbf{0.00E+00} &$\pm$1.38E+01 \\ \hline

\multirow{2}*{\boldsymbol{$F_{10}$}} &9.64E+01 &2.17E+02 &7.92E+02 &3.48E+02 &1.35E+02 &5.08E+01 &\textbf{4.97E+01} &1.06E+02 &9.03E+01  \\ 
&$\pm$1.84E+01 &$\pm$8.69E+01 &$\pm$1.43E+02 &$\pm$9.51E+01 &$\pm$6.67E+02 &$\pm$1.32E+01 &$\pm$\textbf{1.12E+01} &$\pm$1.31E+01 &$\pm$1.79E+01 \\ \hline

\multirow{2}*{\boldsymbol{$F_{11}$}} &1.57E+01 &2.70E+01 &1.89E+01 &2.58E+01 &3.15E+01 &1.68E+01 &\textbf{6.23E+00} &2.53E+01 &1.37E+01  \\ 
&$\pm$1.89E+00 &$\pm$2.18E+00 &$\pm$4.44E+00 &$\pm$4.55E+00 &$\pm$8.45E+00 &$\pm$2.82E+00 &$\pm$\textbf{1.47E+00} &$\pm$1.65E+00 &$\pm$1.27E+00 \\ \hline

\multirow{2}*{\boldsymbol{$F_{12}$}} &7.53E+03 &6.06E+03 &2.28E+03 &1.18E+04 &6.83E+03 &3.11E+03 &1.28E+04 &1.96E+04 &\textbf{1.57E+03}  \\ 
&$\pm$6.72E+03 &$\pm$5.30E+03 &$\pm$3.39E+03 &$\pm$7.82E+03 &$\pm$4.34E+03 &$\pm$2.15E+03 &$\pm$1.53E+04 &$\pm$4.44E+03 &$\pm$\textbf{1.52E+03} \\ \hline

\multirow{2}*{\boldsymbol{$F_{13}$}} &4.32E+00 &1.33E+01 &1.19E+01 &2.80E+01 &7.88E+00 &3.72E+00 &3.35E+00 &\textbf{2.14E+00} &4.54E+00 \\ 
&$\pm$9.03E-01 &$\pm$1.04E+01 &$\pm$3.36E+00 &$\pm$4.44E+00 &$\pm$5.79E+00 &$\pm$5.89E-01 &$\pm$8.52E-01 &$\pm$\textbf{2.09E-01} &$\pm$8.04E-01 \\ \hline

\multirow{2}*{\boldsymbol{$F_{14}$}} &1.34E+01 &1.47E+01 &1.42E+01 &1.35E+01 &1.29E+01 &1.26E+01 &1.47E+01 &1.27E+01 &\textbf{1.24E+01}  \\ 
&$\pm$2.11E-01 &$\pm$1.07E-01 &$\pm$3.11E-01 &$\pm$3.92E-01 &$\pm$3.72E-01 &$\pm$2.71E-01 &$\pm$1.95E-01 &$\pm$2.64E-01 &$\pm$\textbf{3.31E-01} \\ \hline

\multirow{2}*{\boldsymbol{$F_{15}$}} &3.79E+02 &5.72E+02 &8.42E+02 &4.33E+02 &3.00E+02 &3.60E+02 &5.13E+02 &\textbf{6.33E+01} &3.15E+02 \\ 
&$\pm$5.35E+01 &$\pm$1.18E+02 &$\pm$3.19E+02 &$\pm$4.75E+01 &$\pm$7.62E-02 &$\pm$6.51E+01 &$\pm$2.69E+02 &$\pm$\textbf{4.87E+01} &$\pm$5.68E+01 \\ \hline

\multirow{2}*{\boldsymbol{$F_{16}$}} &1.42E+02 &3.77E+02 &5.96E+02 &4.21E+02 &1.44E+02 &\textbf{8.16E+01} &3.39E+02 &1.76E+02 &1.21E+02 \\ 
&$\pm$4.36E+01 &$\pm$1.93E+02 &$\pm$3.35E+02 &$\pm$1.89E+00 &$\pm$7.76E+01 &$\pm$\textbf{6.90E+01} &$\pm$2.99E+02 &$\pm$3.25E+01 &$\pm$1.53E+01 \\ \hline

\multirow{2}*{\boldsymbol{$F_{17}$}} &1.90E+02 &6.46E+02 &8.75E+02 &3.28E+02 &1.58E+02 &\textbf{7.31E+01} &4.15E+02 &2.36E+02 &1.55E+02 \\ 
&$\pm$3.94E+01 &$\pm$3.12E+02 &$\pm$3.34E+02 &$\pm$1.29E+02 &$\pm$7.17E+01 &$\pm$\textbf{2.79E+01} &$\pm$3.07E+02 &$\pm$4.37E+01 &$\pm$2.40E+01 \\ \hline

\multirow{2}*{\boldsymbol{$F_{18}$}} &9.10E+02 &\textbf{8.23E+02} &9.29E+02 &8.32E+02 &9.06E+02 &8.75E+02 &9.04E+02 &9.10E+02 &8.79E+02 \\ 
&$\pm$1.98E+00 &$\pm$\textbf{1.60E+01} &$\pm$1.60E+02 &$\pm$4.00E+01 &$\pm$1.49E+00 &$\pm$6.32E+01 &$\pm$1.86E-01 &$\pm$2.15E+01 &$\pm$8.68E+01 \\ \hline

\multirow{2}*{\boldsymbol{$F_{19}$}} &9.09E+02 &\textbf{8.23E+02} &9.54E+02 &8.45E+02 &9.07E+02 &9.07E+02 &9.25E+02 &9.14E+02 &8.93E+02 \\ 
&$\pm$1.74E+00 &$\pm$\textbf{1.40E+01} &$\pm$1.92E+02 &$\pm$7.77E+01 &$\pm$1.71E+01 &$\pm$4.08E+01 &$\pm$1.07E+02 &$\pm$1.79E+00 &$\pm$4.12E+01 \\ \hline

\multirow{2}*{\boldsymbol{$F_{20}$}} &9.09E+02 &8.29E+02 &1.01E+03 &\textbf{8.24E+02} &9.07E+02 &8.83E+02 &9.04E+02 &9.14E+02 &8.81E+02 \\ 
&$\pm$1.92E+00 &$\pm$3.46E+01 &$\pm$1.95E+02 &$\pm$\textbf{8.86E-01} &$\pm$1.54E+00 &$\pm$5.84E+01 &$\pm$2.32E-01 &$\pm$1.19E+00 &$\pm$1.23E+02 \\ \hline

\multirow{2}*{\boldsymbol{$F_{21}$}} &\textbf{4.96E+02} &8.47E+02 &9.08E+02 &8.22E+02 &5.00E+02 &5.00E+02 &5.12E+02 &5.00E+02 &5.00E+02 \\ 
&$\pm$\textbf{1.81E+01} &$\pm$1.03E+02 &$\pm$3.43E+02 &$\pm$2.60E+02 &$\pm$4.83E-13 &$\pm$2.09E-13 &$\pm$6.00E+01 &$\pm$2.38E-13 &$\pm$2.32E-13 \\ \hline

\multirow{2}*{\boldsymbol{$F_{22}$}} &9.41E+02 &7.45E+02 &1.34E+03 &\textbf{5.74E+02} &9.28E+02 &9.33E+02 &8.24E+02 &9.70E+02 &9.06E+02 \\ 
&$\pm$2.11E+01 &$\pm$2.25E+02 &$\pm$1.60E+02 &$\pm$\textbf{1.27E+02} &$\pm$1.07E+01 &$\pm$2.00E+01 &$\pm$1.59E+01 &$\pm$1.04E+01 &$\pm$1.31E+01 \\ \hline

\multirow{2}*{\boldsymbol{$F_{23}$}} &5.43E+02 &8.36E+02 &1.31E+03 &9.62E+02 &\textbf{5.34E+02} &\textbf{5.34E+02} &5.35E+02 &\textbf{5.34E+02} &5.71E+02 \\ 
&$\pm$1.57E-12 &$\pm$1.13E+02 &$\pm$1.10E+02 &$\pm$3.27E+02 &$\pm$\textbf{4.21E-04} &$\pm$\textbf{2.26E-03} &$\pm$1.88E+00 &$\pm$\textbf{1.57E-04} &$\pm$2.99E+01 \\ \hline

\multirow{2}*{\boldsymbol{$F_{24}$}} &\textbf{2.00E+02} &3.69E+02 &1.57E+03 &2.35E+02 &\textbf{2.00E+02} &\textbf{2.00E+02} &\textbf{2.00E+02} &\textbf{2.00E+02} &\textbf{2.00E+02} \\ 
&$\pm$\textbf{3.59E+02} &$\pm$2.79E+02 &$\pm$1.04E+02 &$\pm$8.36E+01 &$\pm$\textbf{2.96E-09} &$\pm$\textbf{0.00E+00} &$\pm$\textbf{6.39E-13} &$\pm$\textbf{2.67E-12} &$\pm$\textbf{2.72E-12} \\ \hline

\multirow{2}*{\boldsymbol{$F_{25}$}} &1.35E+03 &1.43E+03 &2.00E+03 &1.32E+03 &2.17E+02 &2.13E+02 &2.07E+02 &\textbf{2.00E+02} &2.22E+02 \\ 
&$\pm$3.59E+02 &$\pm$6.85E+01 &$\pm$7.30E+01 &$\pm$3.66E+01 &$\pm$1.59E-01 &$\pm$1.15E+00 &$\pm$6.30E+00 &$\pm$\textbf{1.96E+00} &$\pm$1.37E+01 \\ \hline

\textbf{Friedman-Test} &4.675 &6.150 &7.575 &6.100 &4.850 &3.425 &4.475 &4.600 &\textbf{3.150} \\ \hline

\textbf{Wilcoxon-Test} &\textbf{16-1-3} &\textbf{16-0-4} &\textbf{18-1-1} &\textbf{16-0-4} &\textbf{12-5-3} &\textbf{10-3-7} &\textbf{12-2-6} &\textbf{11-2-7} &--- \\ \hline

\end{tabular}
\end{table*}
\par}

\begin{figure*}[tbp]\renewcommand{\captionfont}{\footnotesize}
  \centering 
  \subfigure[ ]{ 
    \label{fig:subfig:a} 
    \includegraphics[width=1.7in]{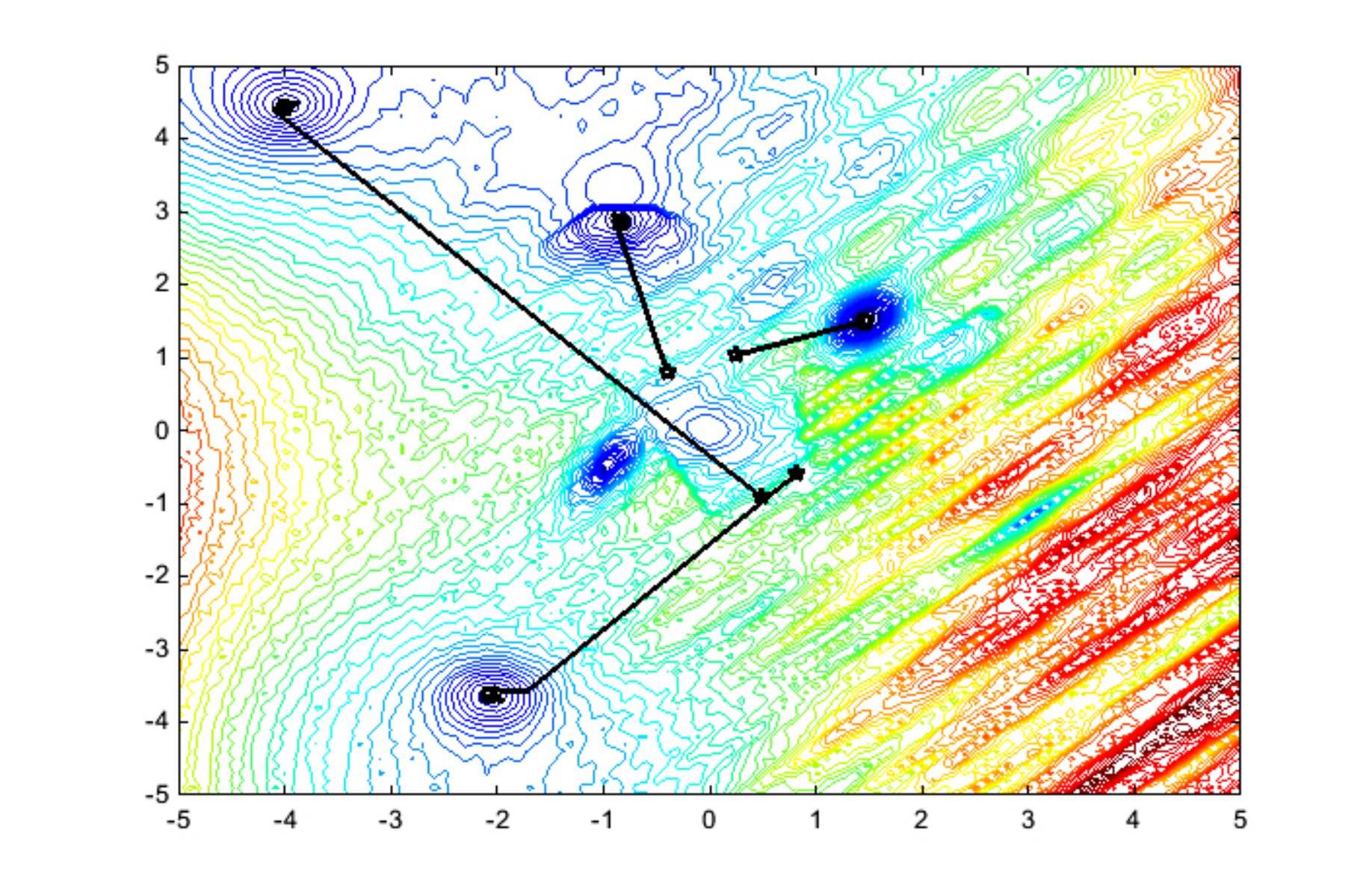}} 
  \subfigure[ ]{ 
    \label{fig:subfig:b} 
    \includegraphics[width=1.7in]{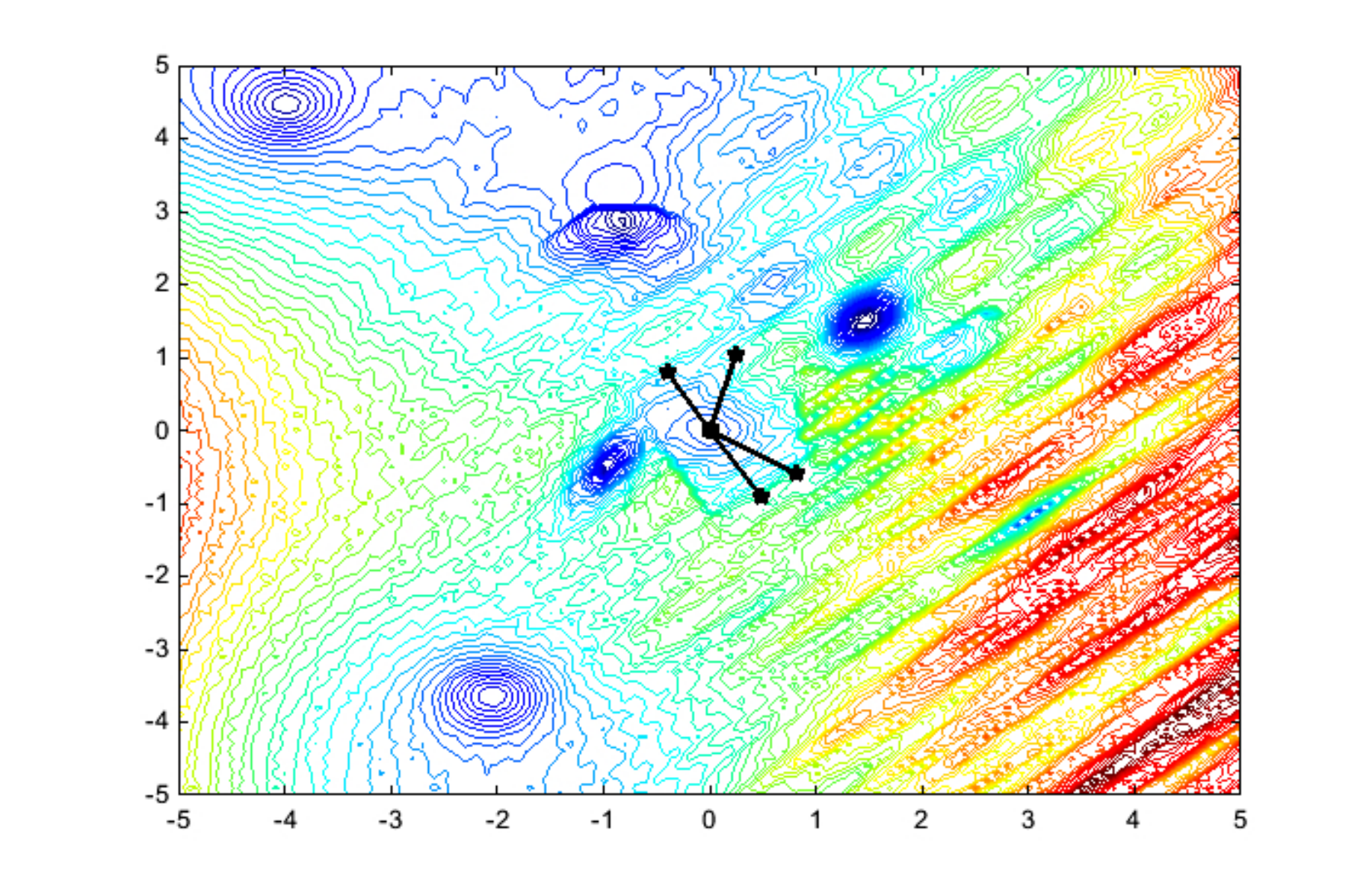}} 
  \label{fig:subfig} 
  \subfigure[ ]{ 
    \label{fig:subfig:c} 
    \includegraphics[width=1.7in]{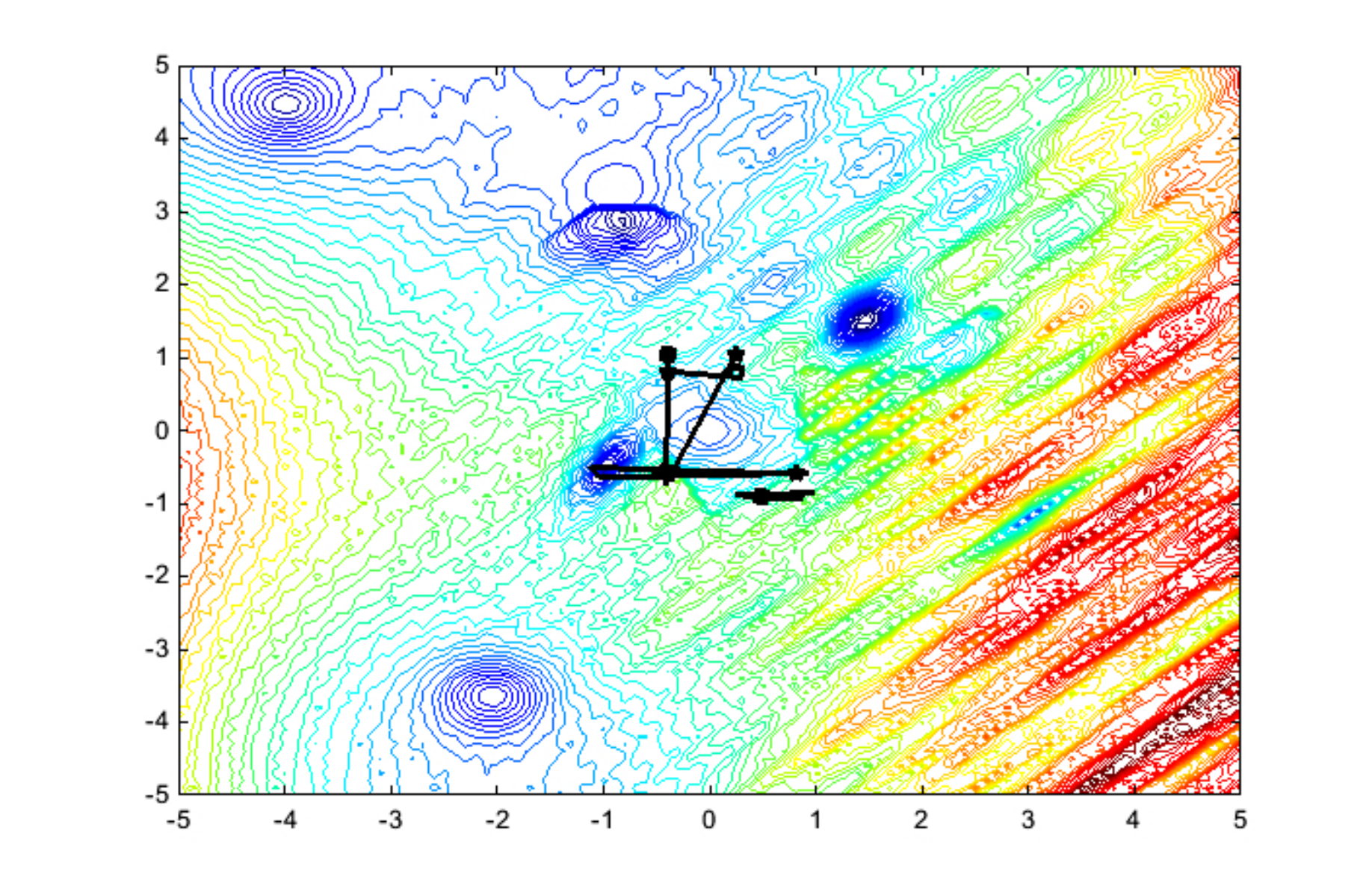}}
  \subfigure[ ]{ 
    \label{fig:subfig:d} 
    \includegraphics[width=1.7in]{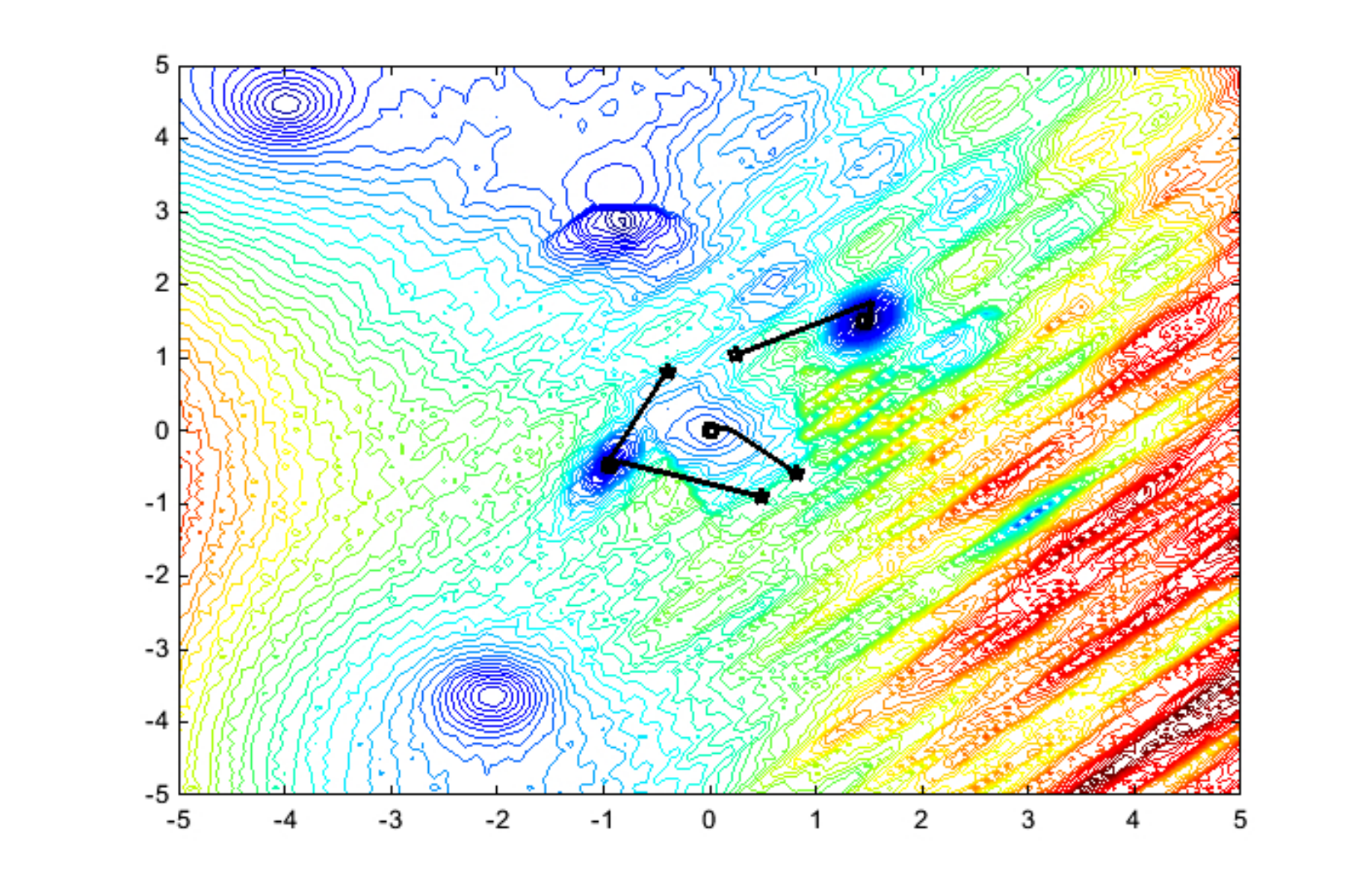}} 
  \caption{The search trajectories of NCS-C (a), SaDE (b), CLPSO (c) and PHC (d) with $N=4$ on the 2-D version of $F_{19}$. The contour lines represent the landscape of the problem. The 4 stars and squares are the locations of initial solutions and final solutions, respectively.} 
  \label{fig:subfig} 
\end{figure*}

\subsection{Algorithm Settings}
To make the comparisons fair, a candidate solution was represented as a real-valued parameter for all the compared algorithms. SA and TS were implemented with the same Gaussian mutation operator, in which the $\sigma_i'$s were initialized to one-tenth of the range of the decision variables. The parameters $r$ and $epoch$ in Eq. (5) were set to 0.99 and 10, respectively. A deterministic cooling schedule was used to control the “Temperature” of SA. Specifically, the temperature was initialized to 1 and then decreased with a factor 0.85 for every 100 iterations \cite{corana1987minimizing}. The TS in \cite{chelouah2000tabu} was used and the 5 most recently generated solutions were kept in the Tabu list. Following its standard procedure \cite{marti2006principles}, the size of ReferenceSet was set 10 for SS and all pairs of solutions in the ReferenceSet were recombined to generate new candidate solutions in each iteration. The 1-point crossover operator was employed as the recombination operator. The Gaussian mutation operator was also applied to each newly generated solution to further improve them. In this experimental study, the standard CMA-ES was compared. GL-25 is a hybrid real-coded genetic algorithm. It first runs the global search during the 25$\%$ of the computational budget, and then executes the local search. The initial population size for the global search is 400, while the local search improves the 100 best solutions output by the global search. In CLPSO, the velocity of a particle is updated by exploiting the personal historical best information of all the particles. In SaDE, both the trial vector generation schemes and the associated control parameters were self-adapted by gradually learning from the previous experiences in generating promising solutions. The population sizes of CLPSO and SaDE were set to 40 and 50, respectively. The PHC copies all the components of NCS-C, except for the calculation of negative correlations between RLSs. In other words, in PHC, the fitness function is regarded as the only criterion when discarding solutions. In addition to $\lambda$, the population size of NCS-C also needs to be predefined. In the experiments, it was set to 10, i.e., the same as SS.

\subsection{Experimental Protocol}
We are mainly interested in multimodal optimization problems since it is this type of problems that motivates the invention of most population-based search methods. The 20 multimodal continuous problems in the benchmark set (numbered as $F_6$-$F_{25}$) for the CEC2005 competition on real-parameter optimization \cite{suganthan2005problem} were used in our empirical studies. The dimensionality of each problem was set to 30. Each compared algorithm terminates when 300 thousands solutions have been generated and the best solution obtained so far was the final solution. The quality of the final solution is measured with function errors, i.e., the difference between the objective function value of the obtained solution and that of the optimal solution to the problem (which are known for these benchmark problems). That is, a better solution corresponds to a smaller function error and a zero function error indicates that the optimal solution is found. Since all the compared algorithms are stochastic search methods, we repeated the experiment for each algorithm for 25 times. The function errors of the corresponding solutions were recorded and the average function errors achieved by the 9 algorithms are presented in Table \uppercase\expandafter{\romannumeral1} together with the standard deviations.

\subsection{Analyses of Results}
The results presented in Table \uppercase\expandafter{\romannumeral1} show that none of the 9 compared algorithms achieved the best results (in terms of average function errors) in all cases. Hence, two-sided Wilcoxon rank-sum statistical tests and the Friedman test have been used to check whether the difference (in terms of function errors) between NCS-C and the compared algorithms are statistically significant. Both  tests were carried out with a 0.05 significance level. By using the Wilcoxon test, we aimed to compare NCS-C with the other algorithms individually, while Friedman test allows us to compare all the 9 algorithms. As shown in the last row of Table \uppercase\expandafter{\romannumeral1}, NCS-C performed statistically significantly better than the other algorithms among the test problems in pairwise comparisons. Furthermore, the Friedman test indicated that NCS-C ranked the highest among the 9 algorithms, as shown in the penultimate row in Table \uppercase\expandafter{\romannumeral1}, and a statistically significant difference was confirmed. 

Furthermore, the number of times an algorithm performed the Kth best, i.e., Top-K (K=1,2,…,9), were counted based on the pairwise two-sided Wilcoxon rank-sum statistical test and depicted in Fig. 1, where an algorithm is the best if its corresponding curve is above those of the other algorithms and reaches 20 (the total number of benchmark problems) with the smallest K. It can be observed that both NCS-C and SaDE first reach 20 at K=6, and the curve of NCS-C is never beneath that of SaDE. Thus, NCS-C is also the best-performing algorithm in this aspect.

Taking a closer look at Table \uppercase\expandafter{\romannumeral1}, NCS-C was the Top-3 algorithm on $F_{6-8}$, $F_{10-12}$, $F_{14-21}$ and $F_{24}$. Among them, $F_6$, $F_{11-12}$, $F_{16-21}$ and $F_{24}$ share the same feature: they consist of a lot of local optima and the local and global optima are located distantly in the search space. A search algorithm is more likely to be trapped in a local optimum on these functions, and may encounter more difficulties to identify better solutions once being trapped. Since NCS-C explicitly encourages multiple RLSs to search different regions of the search space, it is unlikely that all RLSs are trapped in the same local optimum. We believe that this is the main reason that accounts for its good performance on the above problems. This hypothesis can be further supported (albeit indirectly) by two examples. First, $F_{12}$, a variant of the well-known Schwefel function, is known to be a difficult problem for which the landscape consists of several distributed promising areas, where the global optimum may lie in. Thus, exploration is particularly important for this case. The solution quality achieved by NCS-C is much better than those achieved by the other algorithms. The second example is an illustrative one. The NCS-C, SaDE , CLPSO and PHC, all with population size 4, were applied to 2-dimensional versions of the test problems. The trajectories of each RLS/individual were recorded and plotted. The results obtained on $F_{19}$ are presented in Fig. 2. As can be observed from the figure, the RLSs of NCS-C showed negatively correlated search behaviors and each of them finally reached one local (global) optimum exclusively, while SaDE, CLPSO and PHC all prematurely converged and failed to explore the majority of the search space. Similar phenomena have been observed on other test problems. We did not visualize the behavior of SS and GL-25 because they do not maintain a one-to-one mapping between solutions generated in two consecutive iterations.

{\renewcommand\baselinestretch{0.8}\selectfont
\begin{table}[!tbp]\scriptsize
\centering  
\caption{Results of the Antenna Design Experiment. A '-' means results for the corresponding case are not available in the literature.}
\begin{tabular}{c|c|c|c|c|c|c|c} 
\hline
 &  &NCS-C &\cite{lin2010synthesis} &\cite{kurup2003synthesis} &\cite{kumar1999design} &\cite{chen2006modified} &\cite{chen2008application} \\ \hline
\multirow{2}*{\textbf{37-element}} &PO &\textbf{-22.81} &-22.62 &- &-19.37 &-20.49 &-  \\ 
& PP &\textbf{-24.15} &-24.11 &- &- &- &- \\ \hline
\multirow{2}*{\textbf{32-element}} &PO &\textbf{-22.87} &-22.65 &-22.29 &- &- &-  \\ 
& PP &\textbf{-23.71} &-23.45 &-23.22 &- &- &-23.16\\ \hline
\end{tabular}
\end{table}
\par}

\section{Synthesis of Unequally Spaced Antenna Arrays}
To further evaluate NCS on real-world optimization problems in the area of communications, a case study has  been conducted on the SUSAA problem \cite{lin2010synthesis, kurup2003synthesis, kumar1999design, chen2006modified, chen2008application}. Basically, SUSAA can be formulated as a continuous optimization problem that aims at finding an appropriate excitation vector and layout of the elements to generate desirable radiation pattern, i.e., to minimize the Peak Side-Lobe Levels (PSLLs). Following previous works on this problem, the uniform amplitude excitation is taken into account since it can effectively reduce system cost and hardware implementation complexity \cite{lin2010synthesis, kurup2003synthesis, kumar1999design, chen2006modified, chen2008application}. The resultant optimization problem requires identifying the optimal element positions \textbf{x} and excitation phases \boldsymbol{$\phi$} of a linear antenna array that minimizes the PSLL stated in Eq. (9):

\begin{eqnarray} 
\text{PSLL}(\textbf{x},\boldsymbol{\phi})= \underset{\forall{\theta \in S}}{\max}{\vert A_F(\textbf{x},\boldsymbol{\phi},\theta)/A_F(\textbf{x},\boldsymbol{\phi},\theta_0) \vert}
\end{eqnarray}

\noindent where $S$ is the space spanned by angle $\theta$ excluding the mainlobe with the center at $\theta_0$, which was set to $0^\text{o}$. $A_F$(\textbf{x},\boldsymbol{$\phi$},$\theta$) is the array factor of the linear antenna array at angle $\theta$ and can be calculated as

\begin{eqnarray} 
 A_F(\textbf{x},\boldsymbol{\phi},\theta)=\sum_{i=-N}^N{\text{e}^{j(\frac{2\pi x_i\sin(\theta)}{\lambda}+\phi_i)}}
\end{eqnarray}

\noindent where $\lambda$ is wavelength.

The same as [28], the distance between adjacent elements, $d_i=x_i-x_{i-1}$, is restricted to $[0.5\lambda,\lambda]$ for reducing mutual coupling and preventing grating lobes \cite{lin2010synthesis, kurup2003synthesis, kumar1999design, chen2006modified, chen2008application}. The boundary of \boldsymbol{$\phi$} is set to [0, $\pi$]. Besides, a symmetric linear array is considered. Thus, elements satisfy the following constraint:

\begin{eqnarray} 
\left\{ 
\begin{array}{ll}  
 x_{-i}=-x_i \\ 
 \phi_{-i}=\phi_i 
\end{array} 
\right. , \quad\quad
1 \leq i \leq N
\end{eqnarray}

NSC-C is directly applied to the above-described SUSAA problem to re-synthesize a 37-element symmetric linear array and a 32-element symmetric linear array, both of which have been investigated in the literature \cite{lin2010synthesis, kurup2003synthesis, kumar1999design, chen2006modified, chen2008application}. As suggested in \cite{lin2010synthesis, kurup2003synthesis, kumar1999design, chen2006modified, chen2008application}, the angle resolution is set to $0.2^\text{o}$. For each of the four cases, NCS-C repeats 25 runs and for each run, NCS-C stops until the time budget of 500 thousand function evaluations is up.

The average results obtained by NCS-C and the results reported in previous works with exactly the same number of function evaluations are presented in Table \uppercase\expandafter{\romannumeral2}. It can be seen that NCS-C achieved improved performance in all 4 cases. It should be noted that no modification has been made to adapt NCS-C to this specific problem. NSC-C could be further improved with respect to SUSAA by incorporating search mechanisms that suits this problem better. As an illustration, the radiation pattern of the 37-element symmetric array through position-only synthesis is depicted in Fig. 3.

\begin{figure}\renewcommand{\captionfont}{\footnotesize}
			 \centering
			\begin{minipage}[htb]{1\linewidth}
				\centering
				\includegraphics[width=3.5in,height=2.5in]{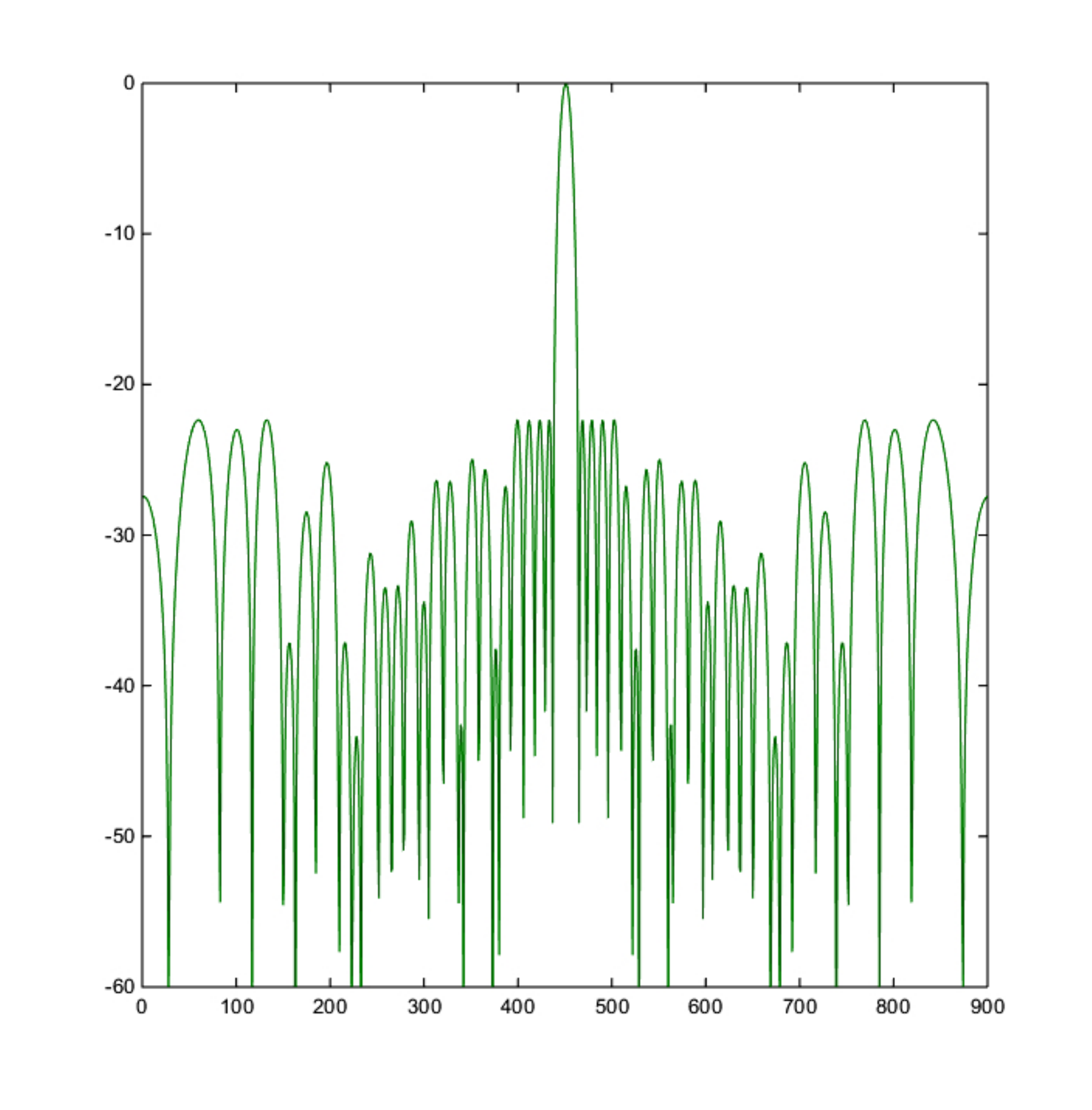}
				\caption{Radiation pattern of 37-element symmetric array through position-only synthesis. The X-axis presents the angle $\theta$, while the Y-axis indicates the $\vert A_F(\textbf{x},\boldsymbol{\phi},\theta)/A_F(\textbf{x},\boldsymbol{\phi},\theta_0) \vert$}
			\end{minipage}%
\end{figure}

\section{Conclusions and Directions for Further Research}

This paper presents a new EA for complex (typically non-convex) optimization problems, which are ubiquitous in communications as well as big data analytics. The proposed approach, namely NCS, is featured by its information sharing and cooperation schemes, which explicitly promote negatively correlated search behaviors in order to explore more effectively in the search space. Empirical results for NCS-C, i.e., an instantiation of NCS for continuous optimization problems, showed the advantages of NCS in comparison to other existing EAs on complex benchmark problems as well as a real-world problem. Further directions for investigations may include: 

\begin{itemize}

\item Theoretical studies of behaviors of NCS, e.g., the time complexity for NCS to achieve the optimal solution.  

\item NCS for set-oriented optimization problems. Set-oriented optimization problems \cite{tantar2013evolve} require seeking a set of solutions that are not only of high quality but also maintain some sort of relationship between them. The simplest version might be to simultaneously find multiple distinct optima for a multimodal problem. Due to its search mechanism, we believe NCS is naturally suitable for this type of problems.

\item Efficient techniques for estimating the Bhattacharyya distance in the context of NCS. The NCS-C employs the Gaussian mutation operator and it is for this reason that the 'correlation' between RLSs could be computed \textit{analytically} and relatively efficiently. Although the probability density function associated with a search operator could be estimated via sampling techniques if it could not be formulated mathematically, efficient techniques for sampling are needed.

\end{itemize}

\bibliographystyle{IEEEtran}
\bibliography{IEEEabrv,NCSrefer}

\begin{IEEEbiography}[{\includegraphics[width=1in,height=1.25in,clip,keepaspectratio]{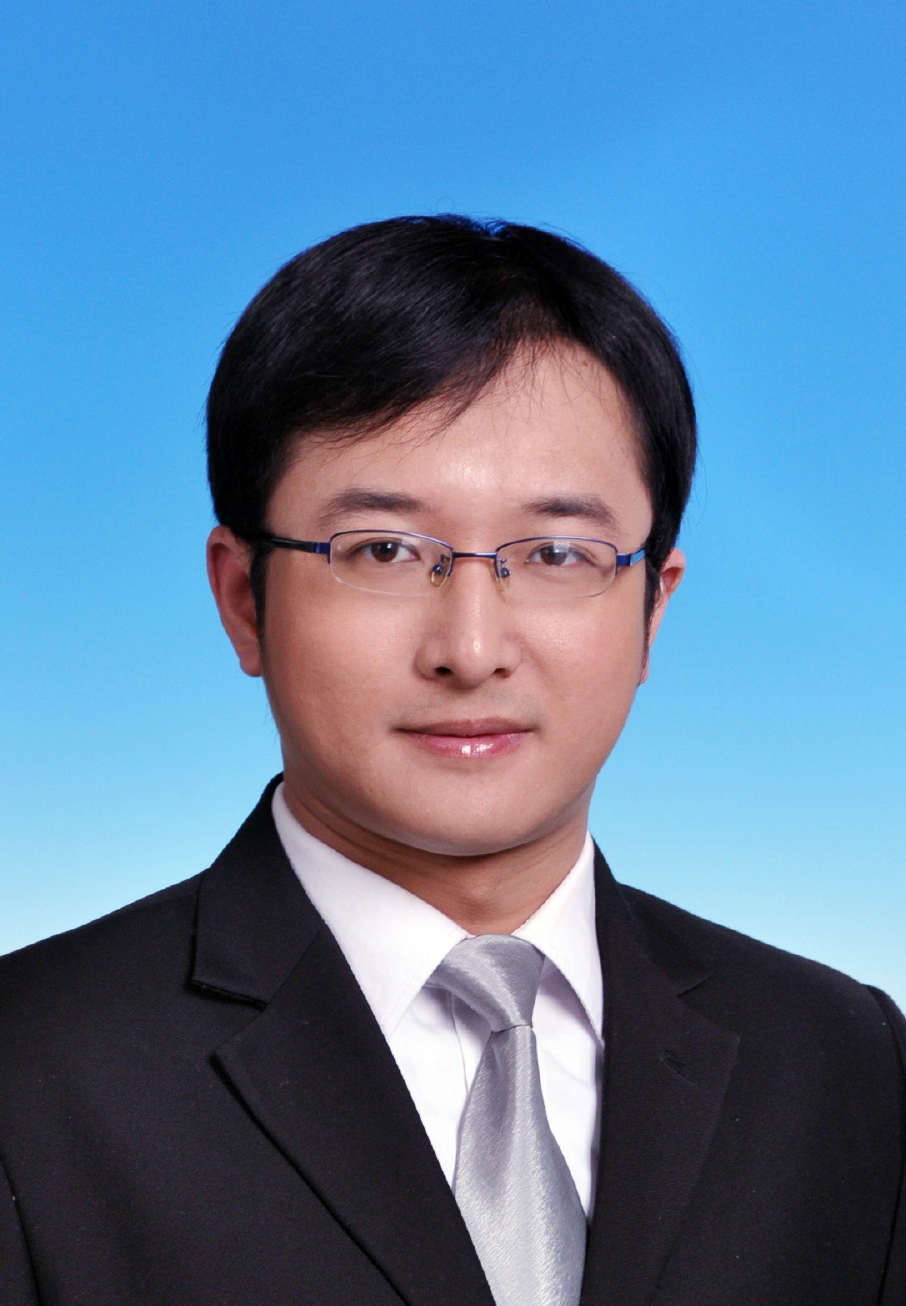}}]{Ke Tang}
\textbf{(S'05-M'07-SM'13)} received the B.Eng. degree from Huazhong University of Science and Technology, Wuhan, China, in 2002, and the Ph.D. degree from Nanyang Technological University, Singapore, in 2007, respectively. 

Since 2007, he has been with the School of Computer Science and Technology, University of Science and Technology of China, where he is currently a Professor. He has authored/co-authored more than 100 refereed publications. His major research interests include evolutionary computation, machine learning, and their real-world applications.

Dr. Tang is an Associate Editor of the IEEE Transactions on Evolutionary Computation, IEEE Computational Intelligence Magazine and Computational Optimization and Applications (Springer), and served as a member of Editorial Boards for a few other journals. He is a member of the IEEE Computational Intelligence Society (CIS) Evolutionary Computation Technical Committee and the IEEE CIS Emergent Technologies Technical Committee. He is the recipient of the Royal Society Newton Advanced Fellowship.

\end{IEEEbiography}

\begin{IEEEbiography}[{\includegraphics[width=1in,height=1.25in,clip,keepaspectratio]{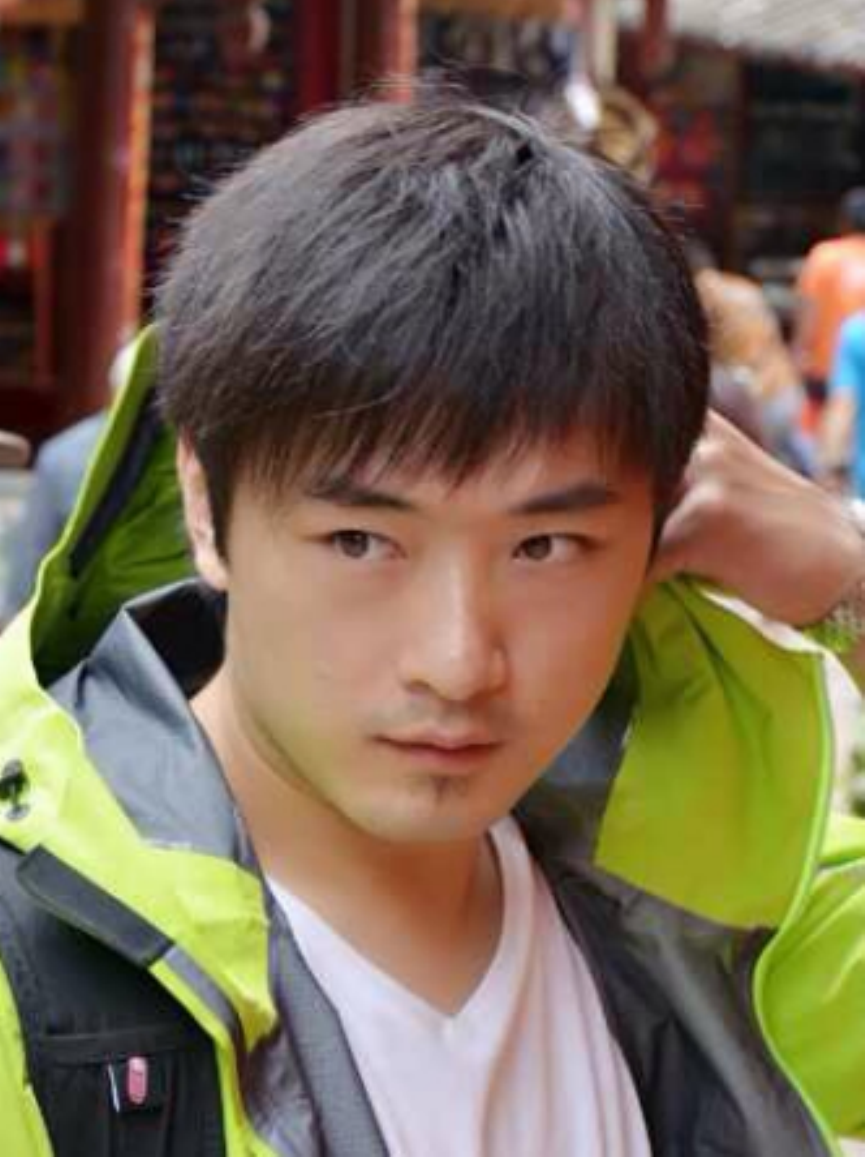}}]{Peng Yang}
\textbf{(S'14)} received his B.Eng. degree in computer science and technology from the University of Science and Technology of China (USTC), Hefei, China, in 2012. He is currently pursuing Ph.D degree from the USTC-Birmingham Joint Research Institute in Intelligent Computation and Its Applications (UBRI), School of Computer Science and Technology, University of Science and Technology of China (USTC). His research interests include evolutionary computation and path planning for unmanned aerial vehicles.
\end{IEEEbiography}

\begin{IEEEbiography}[{\includegraphics[width=1in,height=1.25in,clip,keepaspectratio]{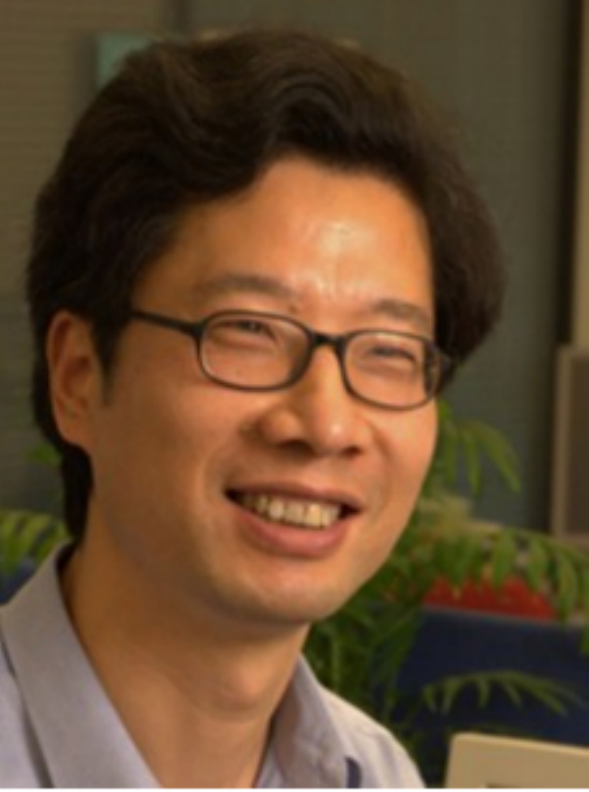}}]{Xin Yao}
\textbf{(F'03)} is a Chair (Professor) of Computer Science and the Director of CERCIA (the Centre of Excellence for Research in Computational Intelligence and Applications) at the University of Birmingham, UK. He is an IEEE Fellow and the President (2014- 15) of IEEE Computational Intelligence Society (CIS).

His major research interests include evolutionary computation and ensemble learning. He published 200+ refereed international journal papers. His papers won the 2001 IEEE Donald G.Fink Prize Paper Award, 2010 and 2015 IEEE Transactions on Evolutionary Computation Outstanding Paper Awards, 2010 BT Gordon Radley Award for Best Author of Innovation (Finalist), 2011 IEEE Transactions on Neural Networks Outstanding Paper Award, and many other best paper awards. He received the prestigious Royal Society Wolfson Research Merit Award in 2012 and the IEEE CIS Evolutionary Computation Pioneer Award in 2013. 

He was the Editor-in-Chief (2003-08) of IEEE Transactions on Evolutionary Computation. 
\end{IEEEbiography}

\end{document}